\documentclass[final]{cvpr}

\usepackage{times}
\usepackage{epsfig}
\usepackage{graphicx}
\usepackage{amsmath}
\usepackage{amssymb}

\usepackage{enumitem}

\usepackage[pagebackref=true,breaklinks=true,colorlinks,bookmarks=false]{hyperref}

\def\cvprPaperID{8} 
\def\confYear{CVPR 2021}

\begin{document}

\title{Temporal Consistency Loss for High Resolution Textured and Clothed 3D Human Reconstruction from Monocular Video}

\author{Akin Caliskan  \hspace{.11\linewidth} Armin Mustafa  \hspace{.11\linewidth} Adrian Hilton\\ \{a.caliskan, armin.mustafa, a.hilton\}@surrey.ac.uk \\
	CVSSP, University of Surrey, UK
	}


\maketitle
\thispagestyle{empty}	
\pagestyle{empty}

\begin{abstract}


We present a novel method to learn temporally consistent 3D reconstruction of clothed people from a monocular video. Recent methods for 3D human reconstruction from monocular video using volumetric, implicit or parametric human shape models, produce per frame reconstructions giving temporally inconsistent output and limited performance when applied to video. In this paper we introduce an approach to learn temporally consistent features for textured reconstruction of clothed 3D human sequences from monocular video by proposing two advances: a novel temporal consistency loss function; and hybrid representation learning for implicit 3D reconstruction from 2D images and  coarse 3D geometry. The proposed advances improve the temporal consistency and accuracy of both the 3D reconstruction and texture prediction from a monocular video. Comprehensive  comparative  performance  evaluation on images of people demonstrates that the proposed method significantly outperforms the state-of-the-art  learning-based single image 3D human shape estimation approaches achieving significant improvement of reconstruction accuracy, completeness, quality and temporal consistency.

\end{abstract}

\section{Introduction}

\begin{figure}[!ht]
	\begin{center}
  \includegraphics[width=\columnwidth]{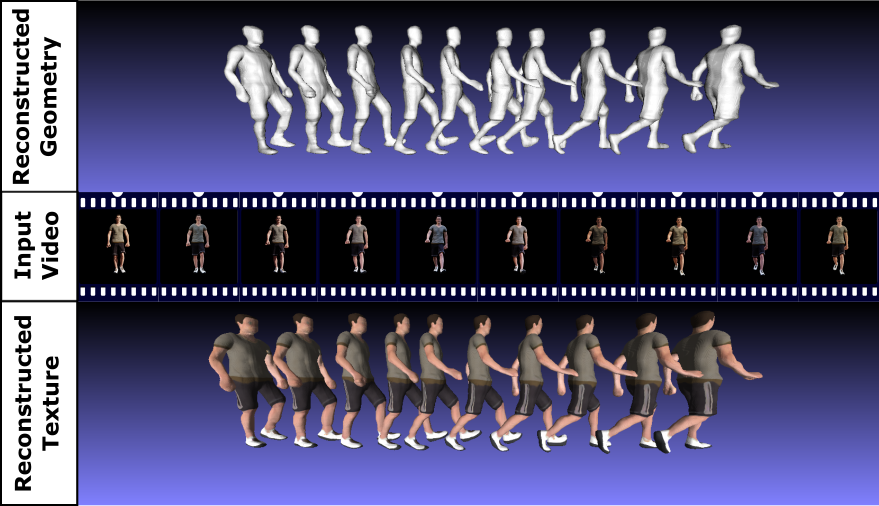}
  \caption{Given an monocular video of a subject (middle), the proposed method creates an accurate and temporally consistent 3D reconstruction (top) with texture (bottom).}
  	\end{center}
  \label{fig:motivation}
\end{figure}

Parsing humans from images is a fundamental task in many applications including AR/VR interfaces \cite{guo2019relightables}, character animation \cite{Mustafa19}, autonomous driving, virtual try-on \cite{dong2019towards} and re-enactment \cite{Liu2018Neural}. There has been significant progress in 2D human pose estimation \cite{cao2017realtime,alp2018densepose}, 2D human segmentation \cite{he2017mask,yang2019parsing} and 3D human pose estimation from monocular video \cite{kocabas2019self,xiang2019monocular,tome2017lifting} to understand the coarse geometry of the human body. 
Recent research has learnt to estimate full 3D human shape from a single image with impressive results \cite{pavlakos2018learning,saito2019pifu,zheng2019deephuman,jackson20183d,varol2018bodynet,caliskan2019learning}. However temporally consistent textured 3D reconstruction of clothed humans from monocular video remains a challenging problem due to the large variation in action, clothing, hair, camera viewpoint, body shape and pose.
This paper addresses this gap in the literature by exploiting a temporally consistent monocular training loss between wide-time separated frames and hybrid implicit-volumetric representation, as shown in Fig. \ref{fig:motivation}.

Traditional multi-view reconstruction methods \cite{Mustafa_2015_ICCV,Leroy_2018_ECCV,gilbert2018volumetric,yu2018doublefusion} have demonstrated the advantages of temporally consistent reconstruction \cite{leroy2017multi,mustafa20174d,mustafa2016temporally}. However, temporally consistent 3D human reconstruction from a monocular RGB video remains an open challenge. Parametric model-based 3D human shape estimation methods have been proposed that exploit temporal neural network architectures \cite{kanazawa2019learning,kocabas2020vibe} for a temporally consistent 3D output. 
Existing parametric models only represent the underlying naked body shape and lack important geometric variation of clothing and hair. Augmented parametric model representations proposed to represent clothing \cite{Bhatnagar2019MultiGarmentNL} are limited to tight clothing which maps bijectively to body shape and do not accurately represent general apparel such as dresses and jackets.

Recent model-free approaches have achieved impressive results in 3D shape reconstruction of clothed people from a single image using learnt volumetric \cite{zheng2019deephuman,jackson20183d,natsume2019siclope,onizuka2020tetratsdf}, point cloud \cite{gabeur2019moulding}, geometry image \cite{pumarola20193dpeople} and implicit \cite{saito2019pifu,saito2020pifuhd} surface representations. 
\cite{caliskan2020multi} proposed multi-view supervision to learn complete and view-consistent 3D human reconstruction. ARCH \cite{huang2020arch} proposed robust 3D reconstruction for arbitrary poses from a single color image and Li et al.\cite{li2020monocular} proposed 3D human reconstruction from video but process frame-by-frame. These methods are trained only with single-image and 3D model pairs without exploiting temporal information between frames. 

We address this problem by proposing a learning framework for textured 3D human reconstruction using temporal consistency for video across wide-timeframes, together with a hybrid 3D volumetric-implicit representation for high resolution textured 3D shape reconstruction.
A volumetric shape representation is learnt using a novel temporal loss function between wide-time separated frames which ensures accurate single-view reconstruction of occluded surface regions. The novel loss function learns to incorporate surface photo-consistency cues in the monocular reconstruction which are not present in the observed image or 3D ground-truth shape. Temporal consistency can only be minimized when the predictions of the trained model are consistent and plausible across all temporal views. The proposed method predicts both high resolution 3D geometry and colored texture from a single view for both visible and unseen human parts.  
The contributions of this work include:
\begin{itemize}[topsep=0pt,partopsep=0pt,itemsep=0pt,parsep=0pt]
\item A novel learning framework for temporally consistent reconstruction of detailed shape and texture for clothed people from monocular video
\item Temporal consistency loss based on wide-timeframe coherence of the shape and appearance reconstruction 
\item A hybrid representation for learning 3D shape which combines the advantages of explicit volumetric representation of occupancy with implicit shape detail
\item The first realistic synthesized video dataset of 400 people with ground-truth 3D models
\end{itemize}

\noindent
The proposed approach learns a temporally consistent hybrid representations giving significant improvement in the accuracy and completeness of reconstruction compared to the state-of-the-art methods for single image human reconstruction \cite{zheng2019deephuman,saito2019pifu,saito2020pifuhd,li2020monocular,caliskan2020multi}.


\begin{figure*}[!ht]
\begin{center}
\includegraphics[width=\textwidth]{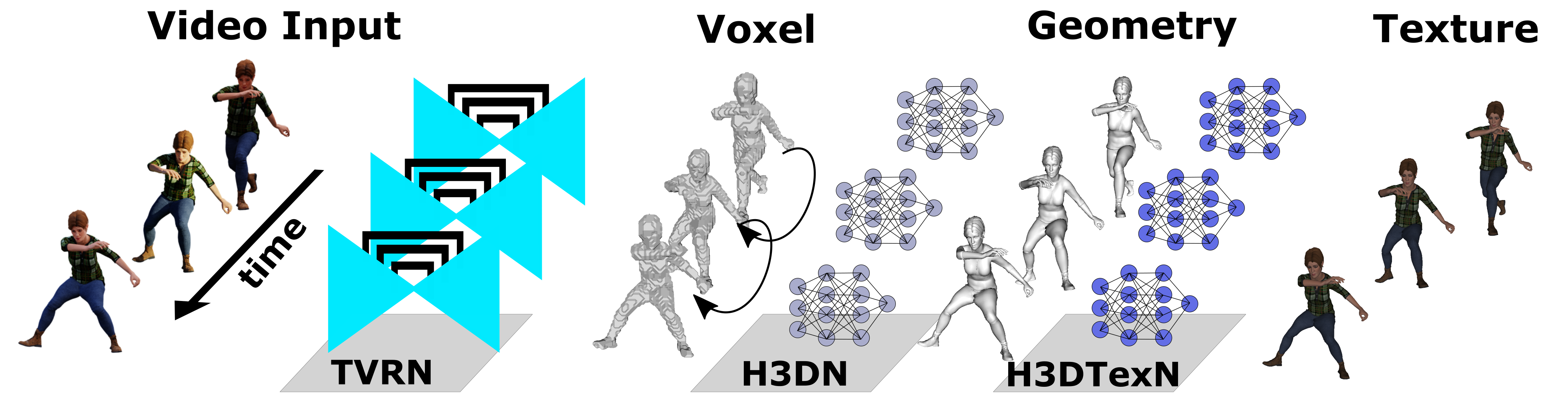}
\caption{The Proposed Framework for Temporally Consistent 3D Human Reconstruction Learning from Video}
\vspace{-0.5cm}
\label{fig:geometry_network}
\end{center}
\end{figure*}
\section{Related Work}
\subsection{Monocular 3D Human Reconstruction}

Parsing 3D humans from a single image can be categorized into model-based and model-free 3D human reconstruction. The first group of approach use parametric human model such as SMPL \cite{SMPL2015,anguelov2005scape} to estimate the body pose and shape parameters in an iterative manner using either 2D joints locations \cite{kanazawa2018end}, 2D joints and silhouettes \cite{bogo2016keep} or 3D joints and mesh coordinates \cite{pavlakos2018learning}. To improve the accuracy of the models, an iterative optimization stage was added to the regression network \cite{kolotouros2019learning}. Even though parametric model-based methods are able to reliably estimate the human body from a single image in the wild, estimated shapes are the naked human body without hair, clothing or other surface details. Recent approaches have extended this to tight-fitting clothing \cite{Ma_2020_CVPR}.

Model-free non-parametric 3D human reconstruction approaches reconstruct clothed people, an overview is given in Table \ref{table:Rel_method_comp}. Model-free methods such as 
Bodynet \cite{varol2018bodynet}, SiCloPe \cite{natsume2019siclope}, DeepHuman \cite{zheng2019deephuman} and MCNet \cite{caliskan2020multi} draw a direct inference of volumetric 3D human shape from a single image. However representing 3D human shape in voxels limits the surface resolution for clothing and hair details. Implicit function networks were introduced to obtain high-resolution 3D reconstruction from a single image. PIFU \cite{saito2019pifu} estimates 3D human reconstruction from a single image  by proposing an implicit decoder which takes pixel-wise image and depth features and predicts occupancy values of 3D points in the encapsulated volume. Following this, PIFUHD \cite{saito2019pifu} improves the previous method by adding features extracted from surface normal maps to the implicit decoder to reconstruct shape detail on the 3D human. However, both methods can not handle large variations in human pose, clothing and hair. ARCH \cite{huang2020arch} proposed a variation of the implicit function network using a parametric model fitted to human silhouettes to improve the 3D reconstruction for arbitrary human poses. Among previous methods, PIFU and ARCH not only reconstruct the 3D geometry of human from a single image, but they also predict the complete texture appearance for the reconstruction. 

Previous methods consider only a single image without any temporal information or consistency. This results in inconsistent reconstruction of shape and appearance when applied to video sequences. We propose a novel method to learn 3D reconstruction of clothed human from a monocular video using image-3D model pairs together with temporal consistency between video frames and 3D models.
\vspace{0.2cm}
\begin{table}[!h] 
        \centering
        \caption{ Comparison of Single View 3D Reconstruction Methods.}
        \label{table:Rel_method_comp}
        \scalebox{0.75}{
        \begin{tabular}{c||c|c|c|c|}
            & \textbf{3D} & \textbf{Temporal} & \textbf{Training} & \textbf{3D Human}  \\
            & \textbf{Represent.} & \textbf{Coherency} & \textbf{Data} & \textbf{Geom. / Text.}  \\ 
            \hline
            Bodynet \cite{varol2018bodynet} & Voxel & No & Image & Yes / No\\ \hline
            SiCloPe \cite{natsume2019siclope} & Voxel & No & Image & Yes / No \\ \hline
            DeepHuman \cite{zheng2019deephuman} & Voxel & No & Image & Yes / No \\ \hline
            3DPeople \cite{pumarola20193dpeople} & Geo. Image & No & Image & Yes / No \\ \hline
            Mould.Hum. \cite{gabeur2019moulding} & Point Cloud & No & Image & Yes / No \\ \hline
            PIFU \cite{saito2019pifu} & Implicit & No & Image & Yes / Yes \\ \hline
            ARCH \cite{huang2020arch} & Implicit & No & Image & Yes / Yes \\ \hline
            PIFU-HD \cite{saito2020pifuhd} & Implicit & No& Image & Yes / No \\ \hline
            MCNet \cite{caliskan2020multi} & Voxel & No & Image & Yes / No \\ \hline
            Geo-PIFU \cite{he2020geo} & Hybrid & No & Image & Yes / No\\ \hline
            \textbf{Proposed} & \textbf{Hybrid} & \textbf{Yes} & \textbf{Video} & \textbf{Yes / Yes}
        \end{tabular}%
        }
\end{table}

\subsection{Temporal Consistency in Neural Networks}
Previous methods in applications other than 3D shape estimation enforce temporal consistency through a \textit{temporal consistency loss} in the context of style transfer \cite{huang2017real}, video-to-video synthesis \cite{wang2018video} or monocular depth estimation \cite{zhang2019exploiting,patil2020don}. 
\textit{Temporal consistency loss} in training or testing encourages similar values along the temporal correspondences estimated from the input video. 
Existing temporally coherent 3D reconstruction methods \cite{leroy2017multi,mustafa20174d,mustafa2016temporally} require multi-view input videos. 
Applying single-image 3D shape estimation methods independently to each frame in a video often produces flickering results. To address this model-based methods have exploited temporal coherency \cite{kanazawa2019learning,kocabas2020vibe}. However, our aim is to predict temporally consistent  model-free 3D clothed human reconstructions from a video, Table \ref{table:Rel_method_comp}. A feed-forward network is used to perform single-view 3D human reconstruction for monocular videos and simultaneously maintain 3D temporal consistency between video frames. Our feed-forward network is trained by enforcing the output 3D reconstruction of temporally distant frames to be both accurate and temporally consistent. Monocular video of a moving person provides significant additional information. The body/clothing shape and appearance of the person should be temporally consistent. 

\subsection{Learning Hybrid 3D Representations}
Model-free single-image 3D human reconstruction methods use various 3D representations-voxel, point cloud, geometry image, and implicit, as seen in Table \ref{table:Rel_method_comp}. Using voxel representation increases computational cost and limits the 3D surface resolution but it keeps global topology and locality of 3D reconstruction. Implicit surface function representation loses the global topology of the 3D human body during inference, but reconstruct the surface with a high level of shape detail. Recent research has combined multiple 3D representations to exploit their relative advantages in the context of 3D deep learning \cite{liu2019point,he2020geo}. \cite{liu2019point} represents the 3D input data in points to reduce the memory consumption, while performing the convolutions in voxels to reduce the irregular, sparse data access and improve the locality. On the other hand, \cite{he2020geo} combine latent voxel features and implicit function learning for 3D geometry prediction. However \cite{chibane2020implicit} shows that using latent features limits ability to learn complex geometry, like articulated shapes, and latent feature approaches do not preserve 3D surface details. Therefore, \cite{chibane2020implicit} proposes an improved 3D shape encoding which is a rich encoding of 3D data through subsequently convolving it with learned convolutions. In the proposed paper, we design a hybrid implicit-volumetric decoder using the shape encoding from temporally coherent 3D voxel reconstruction and image features to predict both 3D geometry and texture of clothed human in an implicit function learning framework. 
%


%

\section{Temporally Consistent 3D Reconstruction}\label{Sec:Method}
This section explains the novel proposed method for temporally consistent textured 3D human reconstruction from a monocular video. An overview of the approach is presented in Fig. \ref{fig:geometry_network}. $N$ frames from a monocular video of a dynamic human with arbitrary pose, clothing, and viewpoint are given as input to the pipeline, and the network predicts the textured 3D human reconstruction in a temporally consistent manner.
%


\begin{figure}[!h]
\begin{center}
\includegraphics[width=\columnwidth]{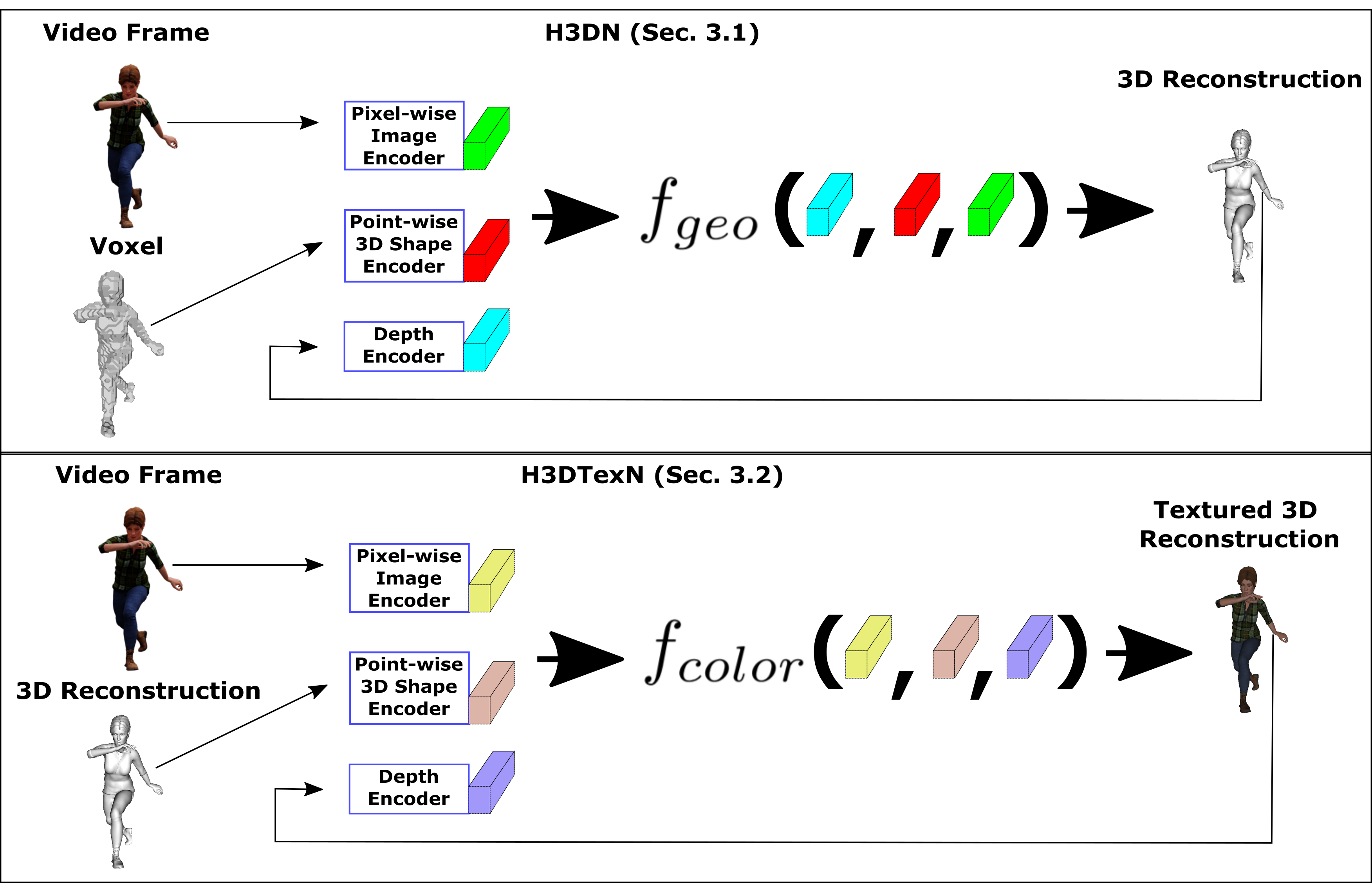}
\caption{This figure shows Hybrid Implicit 3D Reconstruction Network \textit{(H3DN) and Hybrid Implicit 3D Texture Network (\textit{H3dTexN}}). }
\label{fig:geometry_and_texture_network}
\end{center}
\end{figure}

\subsection{Learning 3D Surface Reconstruction}

We propose a method to learn 3D surface reconstruction from a monocular video (Fig. \ref{fig:geometry_network}). In contrast to previous single-view 3D reconstruction approaches, Table \ref{table:Rel_method_comp}, that learn 3D human surface reconstruction from single images, we propose a cascaded network architecture to learn from a monocular video. The proposed architecture consists of \textit{Temporal Voxel Regression Network (TVRN)} and \textit{Hybrid Implicit 3D Reconstruction Network (H3DN)}. \textit{TVRN} reconstructs temporally consistent voxel occupancy grids from a monocular video, and then \textit{H3DN} refines the surface reconstruction. The proposed voxel regression network learns to reconstruct human shape in a temporally consistent manner: $N$ video frames are given to the network such that each frame is passed through its own voxel regression network and parameters are shared between the $N$ networks. The voxel estimation through the TVRN network reduces inconsistency in the reconstruction over time. However, TVRN lacks the high resolution surface details due to the voxel quantisation. Hence the predicted voxel reconstruction is then passed to a hybrid implicit surface function decoder to obtain high-quality surface reconstruction. 


\subsubsection{Learning Architecture}
\label{Sec:Arch}

The proposed learning architecture shown in Fig. \ref{fig:geometry_network} consists of two sub-networks, \textit{Temporal Voxel Regression Network (TRVN)} and \textit{Hybrid Implicit 3D Reconstruction Network (H3DN)}. In previous works, voxel regression has been used to address complete 3D reconstruction of people in a wide variety of poses from a single image \cite{zheng2019deephuman,caliskan2020multi,jackson20183d}. Inspired by these approaches we use a voxel regression network architecture to reconstruct the complete topology of the 3D human from single image. 
To obtain a temporally consistent reconstruction, we introduce a novel learning framework to exploit the temporal consistency between reconstructions from video frames through the proposed \textit{TVRN} architecture (Fig. \ref{fig:geometry_network}).
\textit{TVRN} network consists of multiple parallel stacked hourglass networks with shared parameters. This architecture allows the introduction of a temporal loss function between 3D reconstructions from the input video frames.
As seen in the Fig. \ref{fig:geometry_network}, $N$ frames are used to train the \textit{TVRN} network, including the current frame $I_{t}$ at time $t$ and $N-1$ other frames at different time. The \textit{TVRN} network learns temporally consistent 3D shape and predicts the voxel occupancy grid, $V$ for all frames using the proposed loss function which is computed between predicted voxel occupancy grids for different time frames (Sec. \ref{sec:geo_loss}).


However, the temporally consistent output of \textit{TVRN} has limited surface detail due to the voxel quantisation. To represent high-resolution shape detail we propose \textit{Hybrid Implicit 3D Reconstruction Network (H3DN)} to refine the temporally consistent voxel occupancy grids.
Methods have been proposed in the literature to learn and predict implicit surface representation from a single image \cite{saito2019pifu, saito2020pifuhd, huang2020arch}. These implicit reconstruction methods give a high-level of details on 3D surface. However, all previous methods lose the complete topology of the human body because of the sampling scheme during training. 
In this paper we address this limitation of previous methods by using the voxel occupancy grids as input to the implicit representation instead of a single input image. This allows us to to reconstruct a high-level of surface detail and keep the complete shape topology of the clothed human body due to implicit function learning and input voxel occupancy grids respectively.
The implicit surface is obtained using the proposed novel hybrid implicit function network which takes as input multiple feature encodings and predicts the occupancy of a 3D point. In the network, feature encoders from three different inputs (image, voxel and depth) and a Multi-Layer-Perceptron (MLP) as the decoder to predict occupancy values. 

As illustrated in Fig. \ref{fig:geometry_and_texture_network}, each sampled 3D point ($\mathbf{X}$) is projected on the input image ($\mathbf{x}$) and pixel-wise image features are extracted by concatenating intermediate layer outputs of the hourglass network \cite{saito2019pifu}. We denote the pixel-wise image feature as $\mathcal{H}(I(\mathbf{x}))$. The second input to the decoder are point-wise features extracted from outputs of the \textit{TVRN} network, i.e. voxel occupancy grids. For a sampled 3D point, we apply multi-scale shape encoding \cite{chibane2020implicit} in the aligned voxel occupancy grid using trilinear interpolation within the neighborhood of sampled point. Shape encoding for a 3D sampled point ($\mathbf{X}$) is denoted as $\mathcal{S}(\mathbf{X})$. The last input to the proposed decoder is the depth value of sampled 3D points ($\mathbf{X}$) with respect to the camera, denoted  $\mathcal{D}(\mathbf{X})$. Overall hybrid implicit surface function is formalized as $f_{geometry}$:


\vspace{-0.3cm}
\begin{equation} \label{eq:hybrid_geometry_dec}
f_{geometry}(\mathcal{S}(\mathbf{X}), \mathcal{H}(I(\mathbf{x})), \mathcal{D}(\mathbf{X}) ) = s : s \in \mathbb{R}, s \in [0,1]
\end{equation}

\noindent
The implicit function predicts occupancy values for the sampled 3D points. Marching cubes is applied to obtain a high-quality surface reconstruction.

\subsubsection{Loss Functions}
\label{sec:geo_loss}

\begin{figure}[!ht]
\begin{center}
\includegraphics[width=\columnwidth]{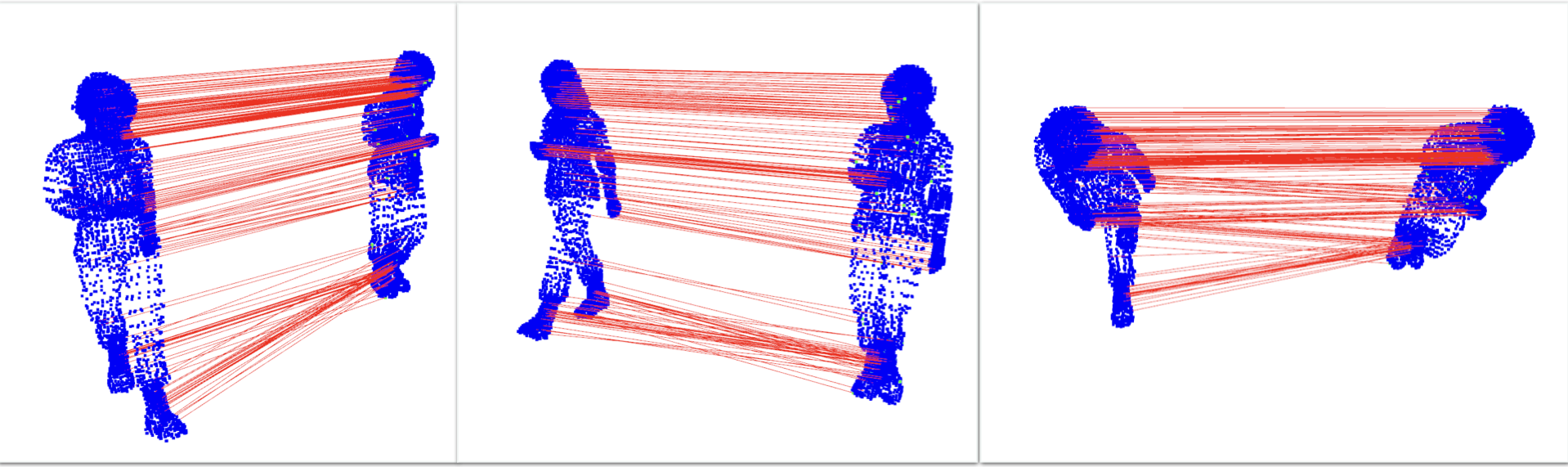}
\caption{This figure shows temporal 3D vertex correspondences between time distant frames from a monocular video to train the proposed networks: \textit{TVRN} and \textit{H3DTexN} (Sec. \ref{Sec:Method})}
\label{fig:temporal_corr}
\end{center}
\end{figure}

The proposed network is supervised from the ground-truth 3D human models rendered from temporal frames and self-supervised between time distant frames from a monocular video. In order to train the \textit{TVRN} network we combine 3D loss $\mathcal{L}_{3D}^{Voxel}$ and temporal consistency loss $\mathcal{L}_{TC}^{Voxel}$.
The 3D loss function $\mathcal{L}_{3D}^{Voxel}$ computes error between the estimated 
3D voxel occupancy grid ($\hat{V}_t$) and 3D ground-truth ($V_t$) for time frame $t$. As stated in Equation \ref{eq:l_3D_1}, the binary cross entropy \cite{jackson2017large} is computed after applying a sigmoid function on the network output. In particular, we used weighted binary cross entropy loss and $\gamma$ is a weight to balance occupied and unoccupied points in the voxel volume:
\vspace{-0.3cm}
\begin{equation} \label{eq:l_3D_1}
\mathcal{L}_{3D}^{Voxel} = { \sum_{t=1}^{N}  \mathcal{L}( \mathcal{V}_{t}, {\hat{\mathcal{V}}_{t}} ) }
\end{equation}
\vspace{-0.3cm}
\begin{eqnarray*} \label{eq:l_3D_2}
\mathcal{L}( \mathcal{V}_{t}, {\hat{\mathcal{V}}_{t}} ) & = &  \sum_{x} \sum_{y} \sum_{z} \gamma\mathcal{V}_{t}^{xyz}\log{{\hat{\mathcal{V}}_{t}}^{xyz}} \nonumber \\  
& + & (1-\gamma)(1-\mathcal{V}_{t}^{xyz})(1-\log{{\hat{\mathcal{V}}_{t}}^{xyz}})  \nonumber
\end{eqnarray*}
\noindent
where ${\mathcal{V}}^{xyz}$ is the occupancy value of a voxel grid ${\mathcal{V}}$ at position $(x,y,z)$. 
Training a network with only binary cross entropy loss gives temporally inconsistent reconstruction for the dynamic parts of the human body as shown in Fig. \ref{fig:geometry_network}. In order to improve 3D model accuracy and completeness, we propose a second loss function, \textit{temporal consistency loss} ($\mathcal{L}_{TC}$) between reconstructions from multiple video frames.
With the \textit{temporal consistency loss}, the representation can learn features robust to temporal changes, self-occlusion and flickering between frames. 3D voxel occupancy grids estimated per frame and the temporal correspondences between vertices are transformed to voxel correspondences as shown in Fig. \ref{fig:temporal_corr}. The \textit{temporal consistency loss} is defined in Equation \ref{eq:l_TC_1}, $L2$ loss is computed between voxel occupancy estimates $\hat{V}$ from one time frame and $N-1$ other frames.
\vspace{-0.3cm}
\begin{equation}\label{eq:l_TC_1}
\mathcal{L}_{TC}^{Voxel} = \sum_{t=1}^{N} \sum_{\substack{l=1\\  l \neq t }}^{N} {\hat{\mathcal{L}}(\hat{\mathcal{V}}_{t}, \hat{\mathcal{V}}_{l} )}
\end{equation}
\vspace{-0.3cm}
\begin{equation*}
\hat{\mathcal{L}}( \hat{\mathcal{V}}_{t}, {\hat{\mathcal{V}}_{l}}) = \sum_{x} \sum_{y} \sum_{z} { \lVert \hat{\mathcal{V}}_{t}^{xyz} - \hat{\mathcal{V}}_{l}^{\mathcal{P}(xyz)} \rVert }_{2}
\end{equation*}
where $\mathbf{\mathcal{P}}$ is the transformation operator between 3D point correspondences. 

In order to train \textit{Hybrid Implicit 3D Reconstruction Network}, we take 3D point samples around the surface of 3D human models and their occupancy values.
To create training point samples, we sample a number $n \in \mathbb{N}$ of points $\mathbf{p}_{i} \in \mathbb{R}^3, i \in 1,\dots,n$ by sampling points on the ground-truth surfaces for every 3D human model, and adding random displacements $\mathbf{n}_i \sim \mathcal{N}(0,\sigma)$, i.e. $\mathbf{p}_{i}^{s} := \mathbf{p}_i + \mathbf{n}_i$. Then we compute the ground truth occupancy values of the sampled points, $o(\mathbf{p}_i^s) \in \{ 0,1\}$. If $\mathbf{p}_{i}^{s}$ is inside the mesh surface, the corresponding $o(\mathbf{p}_i^s) = 1$ and if not, $o(\mathbf{p}_i^s) = 0$. We train the \textit{geometry prediction network} (Equation \ref{eq:hybrid_geometry_dec}) by minimizing the average mean squared error: 
\vspace{-0.3cm}
\begin{equation*}\label{eq:l_HGeo_1}
\mathcal{L}_{3D}^{Hybrid} = \frac{1}{n} \sum_{i=1}^{n} {\lVert f_{geo}(\mathcal{S}(\mathbf{X_i}), \mathcal{H}(I(\mathbf{x_i})), \mathcal{D}(\mathbf{X_i}) ) - o(\mathbf{X_i}) \rVert}_2
\end{equation*}
Both the trained networks are then used to estimate temporally consistent 3D reconstruction from a monocular video. The next section describes estimation of temporally consistent texture for each 3D shape estimation.

\subsection{Learning Textured 3D Reconstruction}

In order to reconstruct complete 3D human models, we propose a temporally consistent texture prediction network in addition to geometry reconstruction, i.e. \textit{Hybrid Implicit 3D Texture Network (H3DTexN)}
The proposed method (Fig. \ref{fig:geometry_and_texture_network}) predicts the color values for each vertex on the reconstructed surface from a monocular video. Different from the previous approaches \cite{saito2019pifu, li2020monocular}, \textit{H3DTexN} learns texture prediction in a temporally consistent manner using the proposed novel hybrid architecture and loss function. 




\subsubsection{Learning Architecture}

The overall method is illustrated in Fig. \ref{fig:geometry_and_texture_network}, which contains $N$ stacked hybrid encoder-decoder networks, where $N$ is number of input video frames to the \textit{H3DTexN} network to learn temporally consistent texture.
The parameters are shared between the hybrid encoder-decoder networks.
Each hybrid encoder-decoder takes multiple feature encodings as input and predicts the RGB color of the sampled 3D point using a Multi-Layer-Perceptron (MLP) decoder to predict color values. 
During training, as illustrated in Fig. \ref{fig:geometry_and_texture_network}, a sampled 3D point ($\mathbf{X}$) is projected on the input image ($\mathbf{x}$) and pixel-wise image feature is extracted by concatenating intermediate layer outputs of a modified Resnet network architecture \cite{johnson2016perceptual}. We denote the pixel-wise image feature $\mathcal{H}(X)$. The second input to the decoder is point-wise features extracted from output of \textit{Hybrid Implicit 3D Reconstruction Network}, i.e. point-wise occupancy values. For a sampled 3D point, we apply multi-scale shape encoding \cite{chibane2020implicit} in the predicted occupancy values within the neighborhood of sampled point using trilinear interpolation. Shape encoding for a sampled point ($\mathbf{X}$) is denoted as $\mathcal{S}(\mathbf{X})$. The last input to the proposed decoder is the depth value of the sampled 3D point ($\mathbf{X}$) with respect to the camera, denoted as $\mathcal{D}(\mathbf{X})$. Overall hybrid implicit texture function is  $f_{color}$:

\vspace{-0.3cm}
\begin{equation} \label{eq:hybrid_color_dec}
f_{color}(\mathcal{S}(\mathbf{X}), \mathcal{H}(I(\mathbf{x})), \mathcal{D}(\mathbf{X}) ) = c : c \in \mathbb{R}^{3\times1}
\end{equation}

The proposed approach learns temporally consistent prediction between different video frames and allows us to use temporal consistency loss (Sec. \ref{sec:text_loss}).  In the overall network, we propose a neural network framework to learn temporally consistent 3D textured human reconstruction from video. The details of the image encoders and MLP networks are explained in  Sec. \ref{Sec:Exp}. In contrast to previous approaches \cite{huang2020arch,saito2019pifu}, the hybrid decoder captures the global topology of the shape with the shape encoding to predict accurate geometry and texture for 3D reconstruction with minimum computational cost.
\subsubsection{Loss Functions}
\label{sec:text_loss}
We train \textit{Hybrid Implicit 3D Texture Network (H3DTexN)} by minimizing two loss functions, $\mathcal{L}_{Color}^{Hybrid}$ and $\mathcal{L}_{Color}^{temporal}$. In order to compute $\mathcal{L}_{Color}^{Hybrid}$, we sample a number $m \in \mathbb{N}$ of points $\mathbf{p}_{i} \in \mathbb{R}^3, i \in 1,\dots,m$ by sampling points on the ground-truth surfaces for every human model. Then, we obtain the color values of the sampled points, $c(\mathbf{p}_i) \in \mathbb{R}^3$. We train the \textit{texture prediction network} by minimizing the average absolute difference error: 
\vspace{-0.3cm}
\begin{equation}\label{eq:l_HText_1}
\mathcal{L}_{Color}^{Hybrid} = \frac{1}{n} \sum_{i=1}^{n} {\left | f_{color}(\mathcal{S}(\mathbf{X_i}), \mathcal{H}(I(\mathbf{x_i})), \mathcal{D}(\mathbf{X_i}) ) - c(\mathbf{X_i}) \right |}
\end{equation}

In order to train the texture prediction network in a temporally consistent manner, we use the temporal correspondences of a sampled point, $\mathbf{p}_{i}^{t}$ for $t = \{1,\dots,N-1$\}, and train the network by minimizing the $L_2$ loss computed between color estimates, $\hat{c}$, from one time frame and $N-1$ other frames:
\vspace{-0.3cm}
\begin{equation}\label{eq:l_HText_2}
\mathcal{L}_{Color}^{Temporal} = \sum_{i=1}^{n} \sum_{\substack{l=1\\  l \neq t }}^{n} { \lVert \hat{c}(\mathbf{X_i}^{t}) - \hat{c}(\mathbf{X_i}^{l}) \rVert}_2
\end{equation}
\begin{equation*}
\hat{c}(\mathbf{X_i})  =  f_{color}(\mathcal{S}(\mathbf{X_i}), \mathcal{H}(I(\mathbf{x_i})), \mathcal{D}(\mathbf{X_i}) ) 
\end{equation*}

The overall loss function, $\mathcal{L}$, is the combined loss function for geometry and color:
\vspace{-0.3cm}
\begin{gather*} \label{eq:l_all}
\mathcal{L} = f(\mathcal{L}_{geo}, \mathcal{L}_{color}) \\
\mathcal{L}_{geo} = h(\mathcal{L}_{3D}^{Voxel}, \mathcal{L}_{TC}^{Voxel}, \mathcal{L}_{3D}^{Hybrid}) \\ 
\mathcal{L}_{color} = g(\mathcal{L}_{Color}^{Hybrid}, \mathcal{L}_{Color}^{Temporal})
\end{gather*}

Testing the proposed method with a monocular video input is illustrated \ref{fig:geometry_network}. Each frame from a monocular video is first passed to \textit{TVRN}. Then, every sample point inside the occupancy volume are given as input to \textit{H3DN} to predict occupancy values for each sample point. Following this, mesh surface reconstruction is estimated from the occupancy volume by applying marching cube algorithm. For appearance, we predict color values of each reconstructed 3D point by using trained \textit{H3DTexN} with the inputs: predicted occupancy volume and a video frame. 




\section{Experimental Evaluation}\label{Sec:Exp}

This section presents the implementation details and synthetic dataset generation together with qualitative and quantitative result on both images and videos of people with varying pose and clothing. We evaluate the proposed method on monocular videos randomly chosen from the datasets. For each video, we give the network video frames and associated segmentation masks. For a given test video, the proposed method estimates the 
temporally consistent surface shape and texture appearance reconstruction based on the framework presented in Section \ref{Sec:Method}.

\begin{figure*}[!t]
\begin{center}
\includegraphics[width=\textwidth]{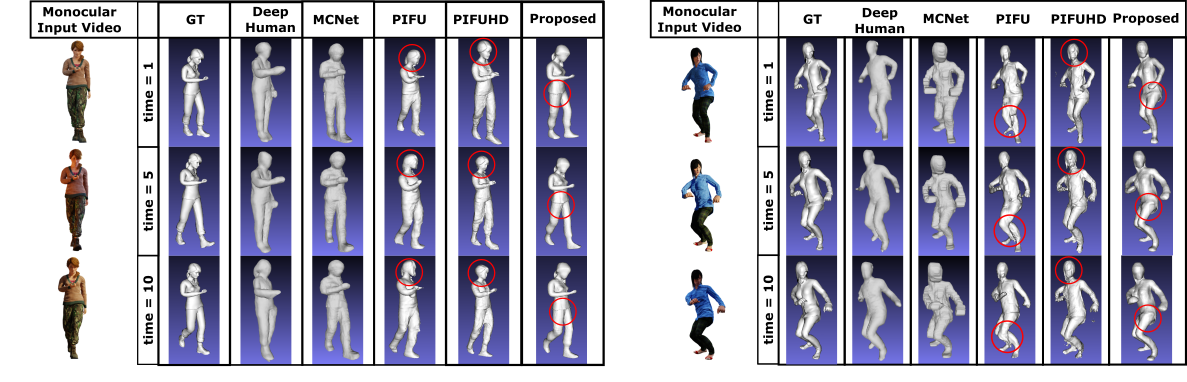}
\caption{Reconstruction results of Deephuman \cite{zheng2019deephuman},  MCNet \cite{caliskan2020multi}, PIFU \cite{saito2019pifu}, PIFUHD \cite{saito2020pifuhd} and the proposed method and ground-truth 3D human models.}
\label{fig:method_comparison_geo_all}
\end{center}
\end{figure*}

\begin{figure}[!h]
\begin{center}
\includegraphics[width=\columnwidth]{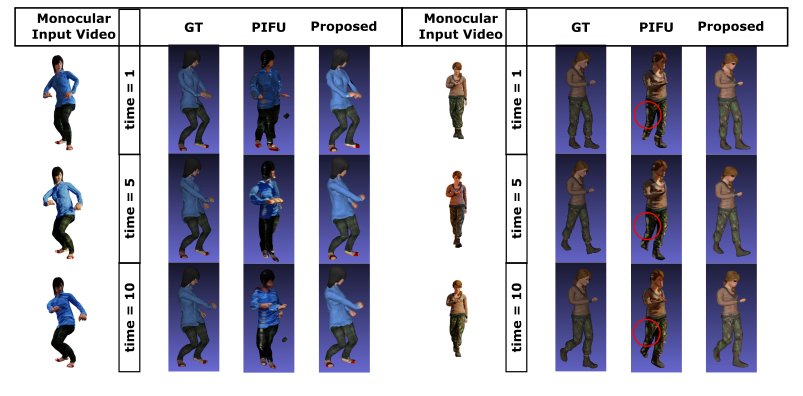}
\caption{Textured reconstruction results of PIFU \cite{saito2019pifu} and the proposed method and ground-truth 3D human models.}
\vspace{-0.2cm}
\label{fig:method_comparison_tex_all}
\end{center}
\end{figure}


\subsection{Datasets}

\begin{figure}[!h]
\begin{center}
\includegraphics[width=\columnwidth]{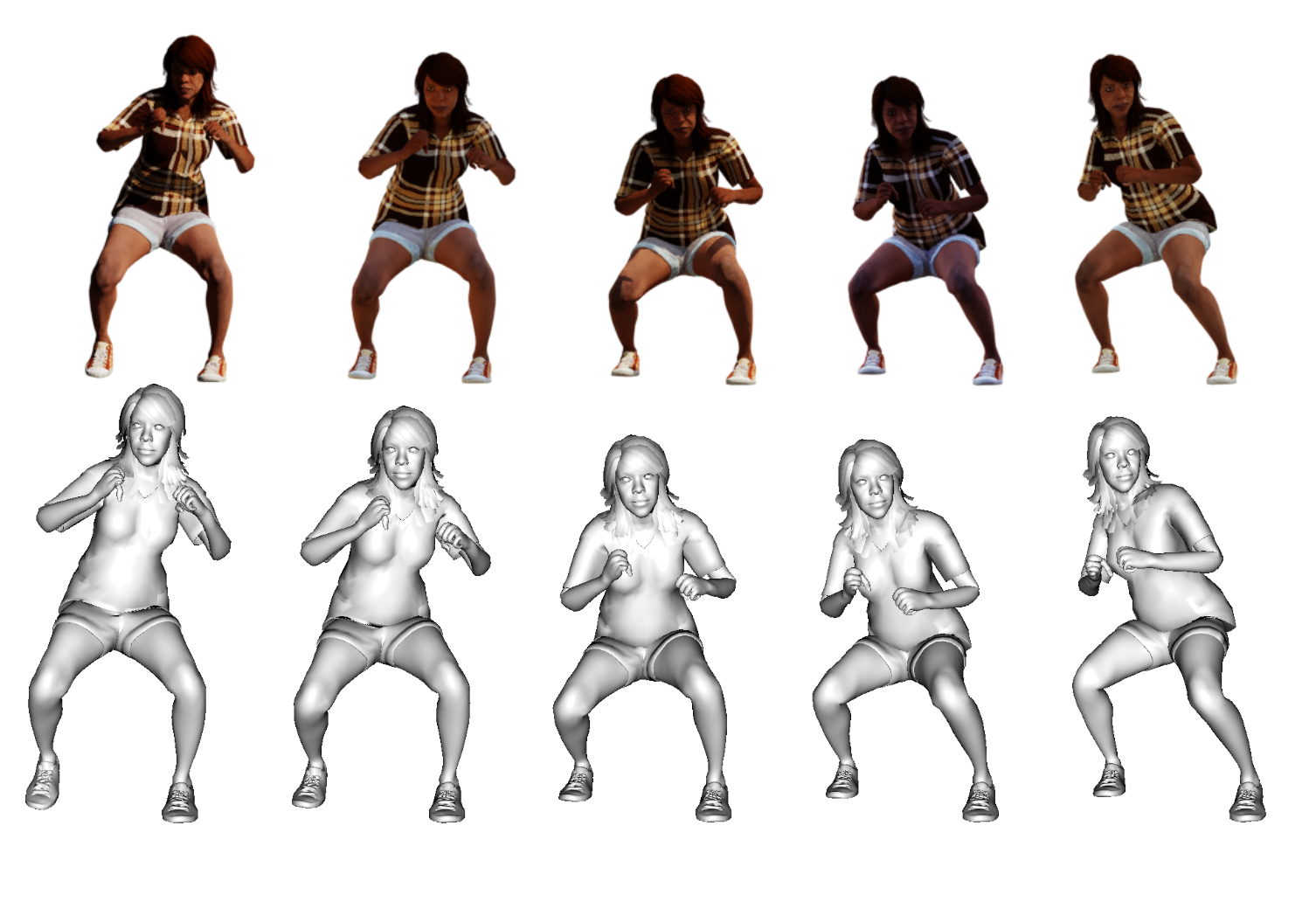}
\caption{This figure shows a sample sequence from \textit{3DVH Video} Dataset.}
\vspace{-0.2cm}
\label{fig:dataset_visual}
\end{center}
\end{figure}

The proposed temporally consistent textured reconstruction from a monocular video is supervised from ground-truth 3D human models and temporal vertex correspondences between video frames. Therefore, we generate a new dataset using similar framework used in the public domain synthetic human image data generation framework \textit{3DVH}\cite{caliskan2020multi} (Fig. \ref{fig:dataset_visual}). Since \textit{3DVH} is limited to static images of 3D humans, we generate 30-frame-length video sequences of the 400 human models with large variations in clothing, hair and pose, which are rendered to 100 camera views per frame. This dataset is referred to as \textit{3DVH Video} and will be made available for research. 

\subsection{Implementation Details}
The proposed network is trained on the \textit{3DVH Video} dataset, which is split into training, validation and test sets. In the \textit{temporal voxel regression network (TVRN)} network the size of the input image is $512\times512\times3$ and output voxel grid resolution is $128\times 128\times 128$. In the ground-truth data the points inside and outside the occupied volume are assigned to 1 and 0 values, respectively. During training, batch size is set to 4 and and epochs to 40. With these settings, the network is trained for 3 days using NVIDIA TitanX with 12GB memory. \textit{TVRN} is trained on relatively low memory GPUs restricting the resolution to $128^3$; however, we can achieve a higher resolution, because \textit{hybrid implicit 3D reconstruction} module can be trained with 3D shape features extracted from higher voxel resolutions. The Adam optimizer is used with learning rate $lr = 2.5e-4$ with the decimation of the step-size every 20 epochs. 

The \textit{hybrid implicit 3D reconstruction (H3DN)} network is trained with the input set of 3D voxels predicted in \textit{TVRN} and video frames. During training of this network, 3D points are sampled around the surface of the 3D ground-truth model (Sec.\ref{sec:geo_loss}). To train the network, we use $10000$ sample points of inside and outside of 3d ground-truth surface. To extract the point-wise shape features from the voxel, we use the 3D convolution architecture from \textit{IFNET} \cite{chibane2020implicit} as it is explained in Sec. \ref{Sec:Arch}. The size of the point-wise features is $[2583\times1]$. For the image encoder, we use an hourglass network architecture \cite{saito2019pifu} to get pixel-wise features of size $[256\times1]$. For the last feature encoder, \textit{depth encoder}, we normalize the actual depth value of the sample point with respect to the camera. In order to predict the occupancy value for the sampled 3D point, concatenation of these features are passed through a Multi Layer Perceptron (MLP) consisting of 5 linear layers of input/output size $[2849, 1024, 512, 256, 128, 1]$, respectively.

For \textit{texture prediction network}, we use the image encoder adapted from CycleGAN architecture \cite{johnson2016perceptual} to extract pixel-wise image features and MLP of 5 linear layers  of input/output size $[2849, 1024, 512, 256, 128, 3]$. 
Both the networks are trained for 5 days using NVIDIA TitanX with 12GB memory, with 2 batch size and 100 epochs. The RMSprop optimizer is used with learning rate $lr = 1e-3$ with the decimation of the step-size every 60 epoch.

\subsection{Evaluation}

\begin{figure}[!h]
\begin{center}
\includegraphics[width=\columnwidth]{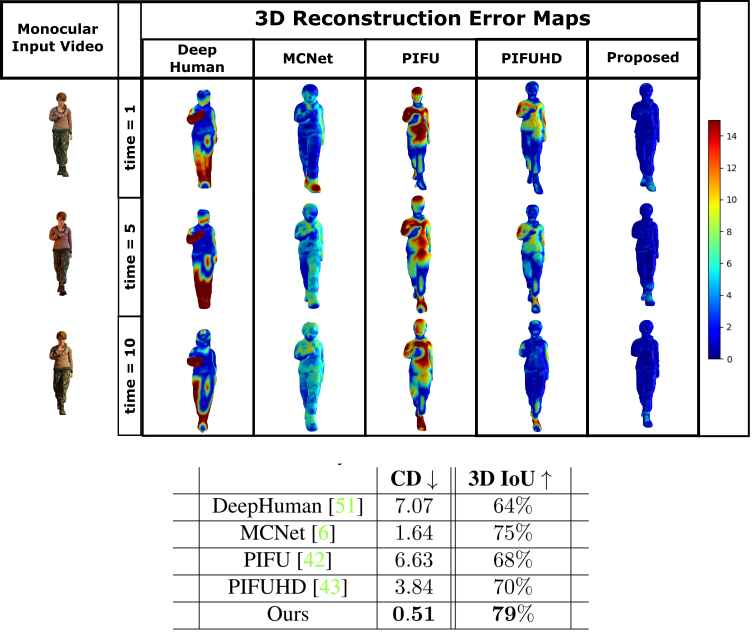}
\caption{\textbf{[Top]} Per vertex chamfer distance from reconstruction to ground-truth model is shown. 3D error maps are illustrated for visible and invisible sides.\textbf{[Bottom]}Comparison of the proposed method with the state-of-the-art methods for different error metrics. \textbf{CD:} Chamfer Distance, \textbf{3D IoU}: 3D Intersection of Union For more details, please refer to the text.}
\label{fig:method_comparison_all_quant}
\end{center}
\end{figure}

The proposed method is qualitatively and quantitatively evaluated against four recent state-of-the-art deep learning-based methods for single image 3D human reconstruction: DeepHuman \cite{zheng2019deephuman}, PIFU \cite{saito2019pifu}, MCNet \cite{caliskan2020multi}, PIFUHD \cite{saito2020pifuhd}. To allow fair comparison, we retrain MCNet, PIFU and Deephuman with the \textit{3DVH Video} dataset using the code provided by the authors and use the pre-trained network of PIFUHD (training code unavailable). 
Qualitative and quantitative comparison of the 3D shape obtained using the proposed approach and the state-of-the-art methods is shown in Fig. \ref{fig:method_comparison_geo_all}, \ref{fig:method_comparison_tex_all} and \ref{fig:method_comparison_all_quant}, along with the ground-truth. All algorithms are tested with monocular video input, and Fig. \ref{fig:method_comparison_geo_all} illustrates the 3D reconstruction results from side views. These results shows that voxel-based methods, DeepHuman and MCNet, are able to predict the coarse 3D reconstruction without cloth and hair details. On the other hand, implicit surface reconstruction methods, PIFU, PIFUHD can reconstruct better surface details while errors occur in the overall topology of the human body for arbitrary poses. For example, Fig. \ref{fig:method_comparison_geo_all} illustrates that PIFU and PIFUHD predict incorrect reconstructions. Also, previous methods shows temporal inconsistency between adjacent reconstructions over time. For example, the PIFU and PIFUHD methods predict inconsistent 3D reconstructions of clothing, face and hair (Fig. \ref{fig:method_comparison_geo_all}). However, the proposed method using a hybrid volumetric-implicit representation trained with a loss-function to enforce temporal consistency results in temporally consistent reconstructions which correctly predicts both body shape and surface details from a single monocular video.

Texture prediction results and comparison of the proposed method with PIFU \cite{saito2019pifu} is shown in Fig. \ref{fig:method_comparison_tex_all}. Textured 3D reconstruction results are shown for both visible and unseen part of the person. PIFU shows limited accuracy for unseen parts while the proposed method is able to predict high-quality temporally consistent texture appearance for the complete surface. In PIFU, the texture prediction network uses image features whereas the proposed method encodes also the shape features into the hybrid implicit surface decoder. The proposed methods also addresses the temporal consistency between frames of the video in Fig. \ref{fig:method_comparison_tex_all}. 

In addition to the qualitative results, we compute two error metrics using the ground-truth 3D models to measure the global accuracy of shape reconstruction: Chamfer Distance (CD) and 3D Intersection of Union (3D IoU) \cite{kaolin2019arxiv}. Fig. \ref{fig:method_comparison_all_quant} shows the comparison of results with ground-truth through the error comparison models with the Chamfer distance error coloured from blue to red as error increases (in centimeters).  Fig. \ref{fig:method_comparison_all_quant} shows the accuracy of reconstructions and temporal consistency between video frames.  The reconstruction obtained using the proposed approach with temporal consistency is significantly better than the 3D shapes obtained using all the previous approaches: DeepHuman, MCNet, PIFU and PIFUHD. In addition to the improvement in the accuracy of the reconstruction, the 3D shape estimated using the proposed method is temporally consistent and the exploitation of temporal redundancy in the learning framework significantly improves the accuracy and completeness of the estimated 3D shape.

\begin{figure}[!h]
\begin{center}
\includegraphics[width=\columnwidth]{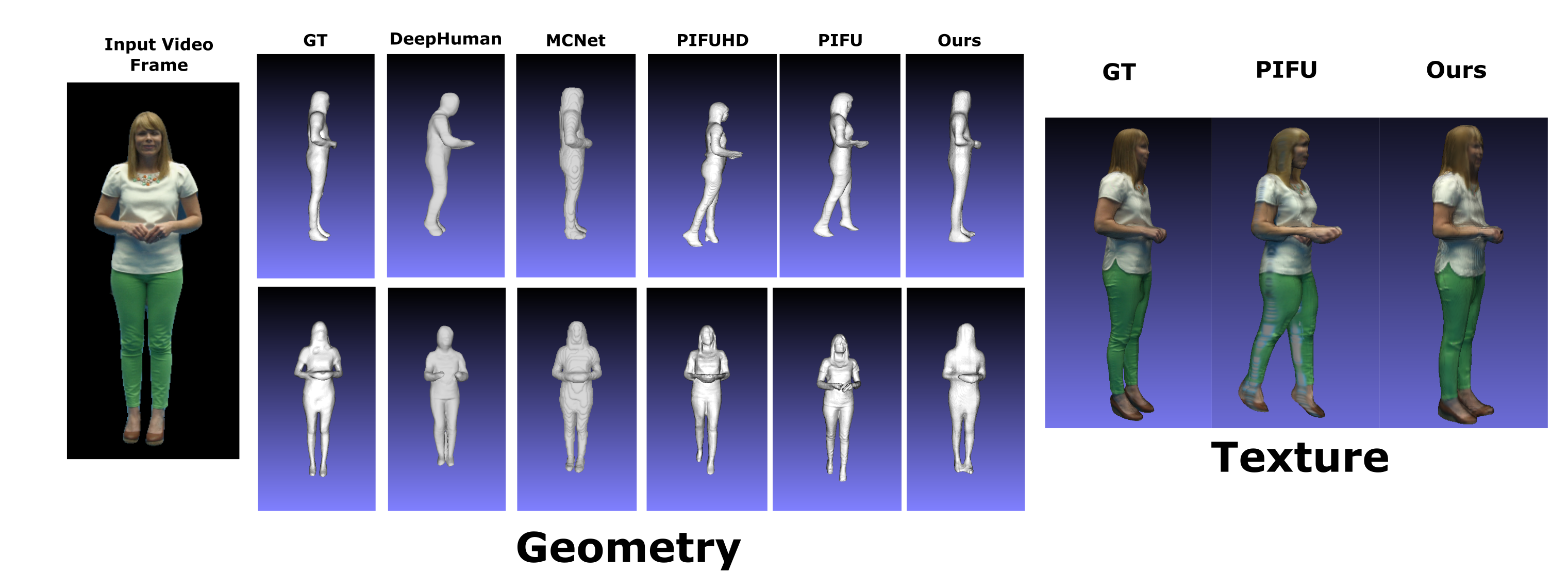}
\caption{Qualitative results on real image from TV presenter dataset \cite{CVSSP_3D}.}
\label{fig:method_comparison_real_all}
\end{center}
\end{figure}

\noindent
\textbf{Real Data Evaluation:}
We evaluate our method against state-of-the-art methods, namely DeepHuman \cite{zheng2019deephuman}, PIFU \cite{saito2019pifu}, MCNet \cite{caliskan2020multi}, PIFUHD \cite{saito2020pifuhd} on the publicly available TV Presenter \cite{CVSSP_3D} dataset, which consists of multiple camera captures of dynamic real humans in a controlled indoor studio. The state-of-the-art methods in figure \ref{fig:method_comparison_real_all} train their models using real datasets: specifically, DeepHuman \cite{zheng2019deephuman} is trained on THuman dataset, and PIFu \cite{saito2019pifu} and PIFuHD \cite{saito2020pifuhd} use the RenderPeople dataset of real human captures. Compared to these methods, the proposed network is trained on the synthetic \textit{3DVH} dataset. 

In figure \ref{fig:method_comparison_real_all}, while DeepHuman \cite{zheng2019deephuman} can recover the coarse shape of human body, the coarse-to-fine approach fails in DeepHuman \cite{zheng2019deephuman} as limbs are missing. Although, PIFu \cite{saito2019pifu} and PIFuHD \cite{saito2020pifuhd} can recover the surface details, they reconstruct legs in wrong position. For textured 3D reconstruction, while PIFu \cite{saito2019pifu} performs reasonable on the visible parts, it fails on unseen part of the human body. Unlike the others, our method is able to recover clothed 3D human body and predict the textured on both visible and unseen parts.

\noindent
\textbf{Limitations:} 
Although the proposed method demonstrates significant improvement in the reconstruction quality over state-of-the-art methods, it suffers from the same limitations as previous methods. The approach assumes complete visibility of the person in the scene and can not handle partial occlusions with objects, as with previous approaches the method also requires silhouettes of the person along with the monocular video for 3D reconstruction.

\section{Conclusion and Future Work}
\vspace{-3pt}
This paper introduces a novel method for temporally consistent textured 3D human reconstruction from a monocular video. The approach is trained with a temporal consistency loss and uses a hybrid volumetric-implicit decoder to learn both overall body shape and surface detail. 
A novel synthetic \textit{3DVH Video} dataset for training is introduced comprising realistic video of 400 people with a wide variation in clothing, hair, body shape, pose and viewpoint. The proposed method demonstrates significant improvement in the reconstruction accuracy, completeness and temporal consistency and improving over state-of-the-art single image methods. Temporal consistency loss together with hybrid implicit decoder are demonstrated to significantly improve the geometry and appearance reconstruction and achieve reliable texture reconstruction of human shape from a monocular video. Future work will exploit self-supervised learning approach for temporally consistent 3D textured human reconstruction from in-the-wild video.

{\small
\bibliographystyle{ieee_fullname}
\bibliography{egbib}

\begin{thebibliography}{10}\itemsep=-1pt

\bibitem{CVSSP_3D}
Multiview video repository,.
\newblock \url{http://cvssp.org/data/cvssp3d/}.
\newblock In Centre for Vision Speech and Signal Processing, University of
  Surrey, UK.

\bibitem{alp2018densepose}
R{\i}za Alp~G{\"u}ler, Natalia Neverova, and Iasonas Kokkinos.
\newblock Densepose: Dense human pose estimation in the wild.
\newblock In {\em Proceedings of the IEEE Conference on Computer Vision and
  Pattern Recognition}, pages 7297--7306, 2018.

\bibitem{anguelov2005scape}
Dragomir Anguelov, Praveen Srinivasan, Daphne Koller, Sebastian Thrun, Jim
  Rodgers, and James Davis.
\newblock Scape: shape completion and animation of people.
\newblock In {\em ACM SIGGRAPH 2005 Papers}, pages 408--416. 2005.

\bibitem{Bhatnagar2019MultiGarmentNL}
Bharat~Lal Bhatnagar, Garvita Tiwari, Christian Theobalt, and Gerard Pons-Moll.
\newblock Multi-garment net: Learning to dress 3d people from images.
\newblock {\em 2019 IEEE/CVF International Conference on Computer Vision
  (ICCV)}, pages 5419--5429, 2019.

\bibitem{bogo2016keep}
Federica Bogo, Angjoo Kanazawa, Christoph Lassner, Peter Gehler, Javier Romero,
  and Michael~J Black.
\newblock Keep it smpl: Automatic estimation of 3d human pose and shape from a
  single image.
\newblock In {\em European Conference on Computer Vision}, pages 561--578.
  Springer, 2016.

\bibitem{caliskan2019learning}
Akin Caliskan, Armin Mustafa, Evren Imre, and Adrian Hilton.
\newblock Learning dense wide baseline stereo matching for people.
\newblock In {\em Proceedings of the IEEE International Conference on Computer
  Vision Workshops}, pages 0--0, 2019.

\bibitem{caliskan2020multi}
Akin Caliskan, Armin Mustafa, Evren Imre, and Adrian Hilton.
\newblock Multi-view consistency loss for improved single-image 3d
  reconstruction of clothed people.
\newblock {\em arXiv preprint arXiv:2009.14162}, 2020.

\bibitem{cao2017realtime}
Zhe Cao, Tomas Simon, Shih-En Wei, and Yaser Sheikh.
\newblock Realtime multi-person 2d pose estimation using part affinity fields.
\newblock In {\em CVPR}, 2017.

\bibitem{chibane2020implicit}
Julian Chibane, Thiemo Alldieck, and Gerard Pons-Moll.
\newblock Implicit functions in feature space for 3d shape reconstruction and
  completion.
\newblock In {\em Proceedings of the IEEE/CVF Conference on Computer Vision and
  Pattern Recognition}, pages 6970--6981, 2020.

\bibitem{dong2019towards}
Haoye Dong, Xiaodan Liang, Xiaohui Shen, Bochao Wang, Hanjiang Lai, Jia Zhu,
  Zhiting Hu, and Jian Yin.
\newblock Towards multi-pose guided virtual try-on network.
\newblock In {\em Proceedings of the IEEE International Conference on Computer
  Vision}, pages 9026--9035, 2019.

\bibitem{gabeur2019moulding}
Valentin Gabeur, Jean-S{\'e}bastien Franco, Xavier Martin, Cordelia Schmid, and
  Gregory Rogez.
\newblock Moulding humans: Non-parametric 3d human shape estimation from single
  images.
\newblock In {\em Proceedings of the IEEE International Conference on Computer
  Vision}, pages 2232--2241, 2019.

\bibitem{gilbert2018volumetric}
Andrew Gilbert, Marco Volino, John Collomosse, and Adrian Hilton.
\newblock Volumetric performance capture from minimal camera viewpoints.
\newblock In {\em Proceedings of the European Conference on Computer Vision
  (ECCV)}, pages 566--581, 2018.

\bibitem{guo2019relightables}
Kaiwen Guo, Peter Lincoln, Philip Davidson, Jay Busch, Xueming Yu, Matt Whalen,
  Geoff Harvey, Sergio Orts-Escolano, Rohit Pandey, Jason Dourgarian, et~al.
\newblock The relightables: volumetric performance capture of humans with
  realistic relighting.
\newblock {\em ACM Transactions on Graphics (TOG)}, 38(6):1--19, 2019.

\bibitem{he2017mask}
Kaiming He, Georgia Gkioxari, Piotr Doll{\'a}r, and Ross Girshick.
\newblock Mask r-cnn.
\newblock In {\em Proceedings of the IEEE international conference on computer
  vision}, pages 2961--2969, 2017.

\bibitem{he2020geo}
Tong He, John Collomosse, Hailin Jin, and Stefano Soatto.
\newblock Geo-pifu: Geometry and pixel aligned implicit functions for
  single-view human reconstruction.
\newblock {\em arXiv preprint arXiv:2006.08072}, 2020.

\bibitem{huang2017real}
Haozhi Huang, Hao Wang, Wenhan Luo, Lin Ma, Wenhao Jiang, Xiaolong Zhu, Zhifeng
  Li, and Wei Liu.
\newblock Real-time neural style transfer for videos.
\newblock In {\em Proceedings of the IEEE Conference on Computer Vision and
  Pattern Recognition}, pages 783--791, 2017.

\bibitem{huang2020arch}
Zeng Huang, Yuanlu Xu, Christoph Lassner, Hao Li, and Tony Tung.
\newblock Arch: Animatable reconstruction of clothed humans.
\newblock In {\em Proceedings of the IEEE/CVF Conference on Computer Vision and
  Pattern Recognition}, pages 3093--3102, 2020.

\bibitem{kaolin2019arxiv}
{Krishna Murthy} J., Edward Smith, Jean-Francois Lafleche, Clement {Fuji
  Tsang}, Artem Rozantsev, Wenzheng Chen, Tommy Xiang, Rev Lebaredian, and
  Sanja Fidler.
\newblock Kaolin: A pytorch library for accelerating 3d deep learning research.
\newblock {\em arXiv:1911.05063}, 2019.

\bibitem{jackson2017large}
Aaron~S Jackson, Adrian Bulat, Vasileios Argyriou, and Georgios Tzimiropoulos.
\newblock Large pose 3d face reconstruction from a single image via direct
  volumetric cnn regression.
\newblock In {\em Proceedings of the IEEE International Conference on Computer
  Vision}, pages 1031--1039, 2017.

\bibitem{jackson20183d}
Aaron~S Jackson, Chris Manafas, and Georgios Tzimiropoulos.
\newblock 3d human body reconstruction from a single image via volumetric
  regression.
\newblock In {\em Proceedings of the European Conference on Computer Vision
  (ECCV)}, pages 0--0, 2018.

\bibitem{johnson2016perceptual}
Justin Johnson, Alexandre Alahi, and Li Fei-Fei.
\newblock Perceptual losses for real-time style transfer and super-resolution.
\newblock In {\em European conference on computer vision}, pages 694--711.
  Springer, 2016.

\bibitem{kanazawa2018end}
Angjoo Kanazawa, Michael~J Black, David~W Jacobs, and Jitendra Malik.
\newblock End-to-end recovery of human shape and pose.
\newblock In {\em Proceedings of the IEEE Conference on Computer Vision and
  Pattern Recognition}, pages 7122--7131, 2018.

\bibitem{kanazawa2019learning}
Angjoo Kanazawa, Jason~Y Zhang, Panna Felsen, and Jitendra Malik.
\newblock Learning 3d human dynamics from video.
\newblock In {\em Proceedings of the IEEE Conference on Computer Vision and
  Pattern Recognition}, pages 5614--5623, 2019.

\bibitem{kocabas2020vibe}
Muhammed Kocabas, Nikos Athanasiou, and Michael~J Black.
\newblock Vibe: Video inference for human body pose and shape estimation.
\newblock In {\em Proceedings of the IEEE/CVF Conference on Computer Vision and
  Pattern Recognition}, pages 5253--5263, 2020.

\bibitem{kocabas2019self}
Muhammed Kocabas, Salih Karagoz, and Emre Akbas.
\newblock Self-supervised learning of 3d human pose using multi-view geometry.
\newblock In {\em Proceedings of the IEEE Conference on Computer Vision and
  Pattern Recognition}, pages 1077--1086, 2019.

\bibitem{kolotouros2019learning}
Nikos Kolotouros, Georgios Pavlakos, Michael~J Black, and Kostas Daniilidis.
\newblock Learning to reconstruct 3d human pose and shape via model-fitting in
  the loop.
\newblock In {\em Proceedings of the IEEE International Conference on Computer
  Vision}, pages 2252--2261, 2019.

\bibitem{leroy2017multi}
Vincent Leroy, Jean-S{\'e}bastien Franco, and Edmond Boyer.
\newblock Multi-view dynamic shape refinement using local temporal integration.
\newblock In {\em Proceedings of the IEEE International Conference on Computer
  Vision}, pages 3094--3103, 2017.

\bibitem{Leroy_2018_ECCV}
Vincent Leroy, Jean-Sebastien Franco, and Edmond Boyer.
\newblock Shape reconstruction using volume sweeping and learned
  photoconsistency.
\newblock In {\em The European Conference on Computer Vision (ECCV)}, September
  2018.

\bibitem{li2020monocular}
Ruilong Li, Yuliang Xiu, Shunsuke Saito, Zeng Huang, Kyle Olszewski, and Hao
  Li.
\newblock Monocular real-time volumetric performance capture.
\newblock {\em arXiv preprint arXiv:2007.13988}, 2020.

\bibitem{Liu2018Neural}
Lingjie Liu, Weipeng Xu, Michael Zollhoefer, Hyeongwoo Kim, Florian Bernard,
  Marc Habermann, Wenping Wang, and Christian Theobalt.
\newblock Neural rendering and reenactment of human actor videos, 2018.

\bibitem{liu2019point}
Zhijian Liu, Haotian Tang, Yujun Lin, and Song Han.
\newblock Point-voxel cnn for efficient 3d deep learning.
\newblock In {\em Advances in Neural Information Processing Systems}, pages
  965--975, 2019.

\bibitem{SMPL2015}
Matthew Loper, Naureen Mahmood, Javier Romero, Gerard Pons-Moll, and Michael~J.
  Black.
\newblock {SMPL}: A skinned multi-person linear model.
\newblock {\em ACM Trans. Graphics (Proc. SIGGRAPH Asia)}, 34(6):248:1--248:16,
  Oct. 2015.

\bibitem{Ma_2020_CVPR}
Qianli Ma, Jinlong Yang, Anurag Ranjan, Sergi Pujades, Gerard Pons-Moll, Siyu
  Tang, and Michael~J. Black.
\newblock Learning to dress 3d people in generative clothing.
\newblock In {\em IEEE/CVF Conference on Computer Vision and Pattern
  Recognition (CVPR)}, June 2020.

\bibitem{Mustafa_2015_ICCV}
Armin Mustafa, Hansung Kim, Jean-Yves Guillemaut, and Adrian Hilton.
\newblock General dynamic scene reconstruction from multiple view video.
\newblock In {\em The IEEE International Conference on Computer Vision (ICCV)},
  December 2015.

\bibitem{mustafa2016temporally}
Armin Mustafa, Hansung Kim, Jean-Yves Guillemaut, and Adrian Hilton.
\newblock Temporally coherent 4d reconstruction of complex dynamic scenes.
\newblock In {\em Proceedings of the IEEE Conference on Computer Vision and
  Pattern Recognition}, pages 4660--4669, 2016.

\bibitem{Mustafa19}
A. Mustafa, C. Russell, and A. Hilton.
\newblock U4d: Unsupervised 4d dynamic scene understanding.
\newblock In {\em ICCV}, 2019.

\bibitem{mustafa20174d}
Armin Mustafa, Marco Volino, Jean-Yves Guillemaut, and Adrian Hilton.
\newblock 4d temporally coherent light-field video.
\newblock In {\em 2017 International Conference on 3D Vision (3DV)}, pages
  29--37. IEEE, 2017.

\bibitem{natsume2019siclope}
Ryota Natsume, Shunsuke Saito, Zeng Huang, Weikai Chen, Chongyang Ma, Hao Li,
  and Shigeo Morishima.
\newblock Siclope: Silhouette-based clothed people.
\newblock In {\em Proceedings of the IEEE Conference on Computer Vision and
  Pattern Recognition}, pages 4480--4490, 2019.

\bibitem{onizuka2020tetratsdf}
Hayato Onizuka, Zehra Hayirci, Diego Thomas, Akihiro Sugimoto, Hideaki
  Uchiyama, and Rin-ichiro Taniguchi.
\newblock Tetratsdf: 3d human reconstruction from a single image with a
  tetrahedral outer shell.
\newblock In {\em Proceedings of the IEEE/CVF Conference on Computer Vision and
  Pattern Recognition}, pages 6011--6020, 2020.

\bibitem{patil2020don}
Vaishakh Patil, Wouter Van~Gansbeke, Dengxin Dai, and Luc Van~Gool.
\newblock Don't forget the past: Recurrent depth estimation from monocular
  video.
\newblock {\em arXiv preprint arXiv:2001.02613}, 2020.

\bibitem{pavlakos2018learning}
Georgios Pavlakos, Luyang Zhu, Xiaowei Zhou, and Kostas Daniilidis.
\newblock Learning to estimate 3d human pose and shape from a single color
  image.
\newblock In {\em Proceedings of the IEEE Conference on Computer Vision and
  Pattern Recognition}, pages 459--468, 2018.

\bibitem{pumarola20193dpeople}
Albert Pumarola, Jordi Sanchez, Gary Choi, Alberto Sanfeliu, and Francesc
  Moreno-Noguer.
\newblock {3DPeople: Modeling the Geometry of Dressed Humans}.
\newblock In {\em International Conference on Computer Vision (ICCV)}, 2019.

\bibitem{saito2019pifu}
Shunsuke Saito, Zeng Huang, Ryota Natsume, Shigeo Morishima, Angjoo Kanazawa,
  and Hao Li.
\newblock Pifu: Pixel-aligned implicit function for high-resolution clothed
  human digitization.
\newblock In {\em Proceedings of the IEEE International Conference on Computer
  Vision}, pages 2304--2314, 2019.

\bibitem{saito2020pifuhd}
Shunsuke Saito, Tomas Simon, Jason Saragih, and Hanbyul Joo.
\newblock Pifuhd: Multi-level pixel-aligned implicit function for
  high-resolution 3d human digitization.
\newblock In {\em Proceedings of the IEEE/CVF Conference on Computer Vision and
  Pattern Recognition}, pages 84--93, 2020.

\bibitem{tome2017lifting}
Denis Tome, Chris Russell, and Lourdes Agapito.
\newblock Lifting from the deep: Convolutional 3d pose estimation from a single
  image.
\newblock In {\em Proceedings of the IEEE Conference on Computer Vision and
  Pattern Recognition}, pages 2500--2509, 2017.

\bibitem{varol2018bodynet}
Gul Varol, Duygu Ceylan, Bryan Russell, Jimei Yang, Ersin Yumer, Ivan Laptev,
  and Cordelia Schmid.
\newblock Bodynet: Volumetric inference of 3d human body shapes.
\newblock In {\em Proceedings of the European Conference on Computer Vision
  (ECCV)}, pages 20--36, 2018.

\bibitem{wang2018video}
Ting-Chun Wang, Ming-Yu Liu, Jun-Yan Zhu, Guilin Liu, Andrew Tao, Jan Kautz,
  and Bryan Catanzaro.
\newblock Video-to-video synthesis.
\newblock {\em arXiv preprint arXiv:1808.06601}, 2018.

\bibitem{xiang2019monocular}
Donglai Xiang, Hanbyul Joo, and Yaser Sheikh.
\newblock Monocular total capture: Posing face, body, and hands in the wild.
\newblock In {\em Proceedings of the IEEE Conference on Computer Vision and
  Pattern Recognition}, pages 10965--10974, 2019.

\bibitem{yang2019parsing}
Lu Yang, Qing Song, Zhihui Wang, and Ming Jiang.
\newblock Parsing r-cnn for instance-level human analysis.
\newblock In {\em Proceedings of the IEEE Conference on Computer Vision and
  Pattern Recognition}, pages 364--373, 2019.

\bibitem{yu2018doublefusion}
Tao Yu, Zerong Zheng, Kaiwen Guo, Jianhui Zhao, Qionghai Dai, Hao Li, Gerard
  Pons-Moll, and Yebin Liu.
\newblock Doublefusion: Real-time capture of human performances with inner body
  shapes from a single depth sensor.
\newblock In {\em Proceedings of the IEEE conference on computer vision and
  pattern recognition}, pages 7287--7296, 2018.

\bibitem{zhang2019exploiting}
Haokui Zhang, Chunhua Shen, Ying Li, Yuanzhouhan Cao, Yu Liu, and Youliang Yan.
\newblock Exploiting temporal consistency for real-time video depth estimation.
\newblock In {\em Proceedings of the IEEE International Conference on Computer
  Vision}, pages 1725--1734, 2019.

\bibitem{zheng2019deephuman}
Zerong Zheng, Tao Yu, Yixuan Wei, Qionghai Dai, and Yebin Liu.
\newblock Deephuman: 3d human reconstruction from a single image.
\newblock In {\em Proceedings of the IEEE International Conference on Computer
  Vision}, pages 7739--7749, 2019.

\end{thebibliography}
}

\end{document}